\documentclass[sigconf]{acmart}

\usepackage{algorithm}
\usepackage{algorithmic}

\usepackage{microtype}
\usepackage{amsmath}
\usepackage{amsthm}
\usepackage{graphicx}
\usepackage{booktabs}
\usepackage{multirow}

\usepackage{array}
\usepackage{arydshln}
\usepackage{color}
\newenvironment{enumeratesquish}[2]{\begin{list}{\labelenumi}{\setlength{\itemsep}{#1}\setlength{\labelwidth}{#2}\setlength{\leftmargin}{\labelwidth}\addtolength{\leftmargin}{\labelsep}}}{\end{list}}

\AtBeginDocument{%
  }

\copyrightyear{2024}
\acmYear{2024}
\setcopyright{rightsretained}
\acmConference[CIKM '24]{Proceedings of the 33rd ACM International Conference on Information and Knowledge Management}{October 21--25, 2024}{Boise, ID, USA} \acmBooktitle{Proceedings of the 33rd ACM International Conference on Information and Knowledge Management (CIKM '24), October 21--25, 2024, Boise, ID, USA} \acmDOI{10.1145/3627673.3679653} \acmISBN{979-8-4007-0436-9/24/10}

\makeatletter
\gdef\@copyrightpermission{
  \begin{minipage}{0.3\columnwidth}
   \href{https://eur02.safelinks.protection.outlook.com/?url=https\%3A\%2F\%2Fcreativecommons.org\%2Flicenses\%2Fby-nc-sa\%2F4.0\%2F&data=05\%7C02\%7C\%7Ccdfabbd64801477933c308dcc07ee709\%7C2e9f06b016694589878910a06934dc61\%7C0\%7C0\%7C638596898490442569\%7CUnknown\%7CTWFpbGZsb3d8eyJWIjoiMC4wLjAwMDAiLCJQIjoiV2luMzIiLCJBTiI6Ik1haWwiLCJXVCI6Mn0\%3D\%7C0\%7C\%7C\%7C&sdata=U5A9HEgy2wCStELw8J4U565t9u7kOAySuVhMPhoQlAE\%3D&reserved=0}{\includegraphics[width=0.90\textwidth]{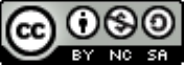}}
  \end{minipage}\hfill
  \begin{minipage}{0.7\columnwidth}
   \href{https://eur02.safelinks.protection.outlook.com/?url=https\%3A\%2F\%2Fcreativecommons.org\%2Flicenses\%2Fby-nc-sa\%2F4.0\%2F&data=05\%7C02\%7C\%7Ccdfabbd64801477933c308dcc07ee709\%7C2e9f06b016694589878910a06934dc61\%7C0\%7C0\%7C638596898490446536\%7CUnknown\%7CTWFpbGZsb3d8eyJWIjoiMC4wLjAwMDAiLCJQIjoiV2luMzIiLCJBTiI6Ik1haWwiLCJXVCI6Mn0\%3D\%7C0\%7C\%7C\%7C&sdata=TVxBGDQfxFGOnLr36yedsNci8F5szQjGD5I6z0Kfaa0\%3D&reserved=0}{This work is licensed under a Creative Commons Attribution-NonCommercial-ShareAlike International 4.0 License.}
  \end{minipage}
  \vspace{5pt}
}
\makeatother

\begin{document}

\title{MANA-Net: Mitigating Aggregated Sentiment Homogenization with News Weighting for Enhanced Market Prediction}
\author{Mengyu Wang}
\orcid{0009-0002-4160-3791}
\affiliation{%
  \institution{The University of Edinburgh}
  \city{Edinburgh}
  \country{UK}
}
\email{mengyu.wang@ed.ac.uk}

\author{Tiejun Ma}
\orcid{0000-0001-5545-6978}
\affiliation{
  \institution{The University of Edinburgh}
  \city{Edinburgh}
  \country{UK}
}
\email{tiejun.ma@ed.ac.uk}

\renewcommand{\shortauthors}{Mengyu Wang and Tiejun Ma}

\begin{abstract}
It is widely acknowledged that extracting market sentiments from news data benefits market predictions. However, existing methods of using financial sentiments remain simplistic, relying on equal-weight and static aggregation to manage sentiments from multiple news items. This leads to a critical issue termed ``Aggregated Sentiment Homogenization'', which has been explored through our analysis of a large financial news dataset from industry practice. This phenomenon occurs when aggregating numerous sentiments, causing representations to converge towards the mean values of sentiment distributions and thereby smoothing out unique and important information. Consequently, the aggregated sentiment representations lose much predictive value of news data. 
To address this problem, we introduce the Market Attention-weighted News Aggregation Network (MANA-Net), a novel method that leverages a dynamic market-news attention mechanism to aggregate news sentiments for market prediction. MANA-Net learns the relevance of news sentiments to price changes and assigns varying weights to individual news items. By integrating the news aggregation step into the networks for market prediction, MANA-Net allows for trainable sentiment representations that are optimized directly for prediction. We evaluate MANA-Net using the S\&P 500 and NASDAQ 100 indices, along with financial news spanning from 2003 to 2018. Experimental results demonstrate that MANA-Net outperforms various recent market prediction methods, enhancing Profit \& Loss by 1.1\% and the daily Sharpe ratio by 0.252.
\end{abstract}



\maketitle

\section{Introduction}
\label{sec:introduction}

Many studies have documented the significant influence of news on financial markets~\cite{goonatilake2007volatility, barber2008all}. News sentiment analysis has emerged as a prominent research area, aiming to capture the emotional tone of news articles and their potential impact on market movements~\cite{loughran2011liability}. Unlike general sentiment analysis, financial news sentiment analysis deals with specialized vocabulary and requires consideration of market context~\cite{mishev2020evaluation}. Advancements in Natural Language Processing (NLP) have led to the development of tools for financial sentiment extraction, including dictionary-based methods~\cite{dewally2003internet}, deep learning methods~\cite{araci2019finbert, wu2021equity2vec}, and recent large language models (LLMs)~\cite{lopez2023can}.

Despite the diverse methodologies for financial news sentiment analysis, the utilization of sentiments from multiple news items remains simplistic. Recent news-based stock prediction works, including various multi-modal models~\cite{wang2018combining, mohan2019stock, li2020multimodal}, and LLMs~\cite{lopez2023can}, primarily focus on prediction methods themselves. These methods commonly employ aggregation techniques like sentiment averaging, polarity counting, or custom sentiment indicators to derive variables representing overall market sentiments. However, these aggregation approaches are equal-weight and static, meaning they treat all news as equally influential and consistently generate fixed aggregated representations. This fails to capture the dynamic nature of market sentiments, where news articles exert varying degrees of influence depending on market conditions~\cite{clapham2021popular, cheung2019exchange, sawhney2021fast}. Additionally, market efficiency theory suggests that key news information is reflected in market prices~\cite{fama1970efficient, zhang2016market}. However, static aggregation does not account for the interactions between prices and news. Consequently, these equal-weight and static methods may not fully extract the predictive power available in news data, especially when dealing with large volumes of news.

\begin{figure}[h]
\centering
\vspace{-0.8cm}
\begin{center}
   \captionof{figure}{The kernel density estimate plot illustrates the distributions of sentiment scores for both individual news items and their daily average values.}
   \includegraphics[width=0.98\linewidth]{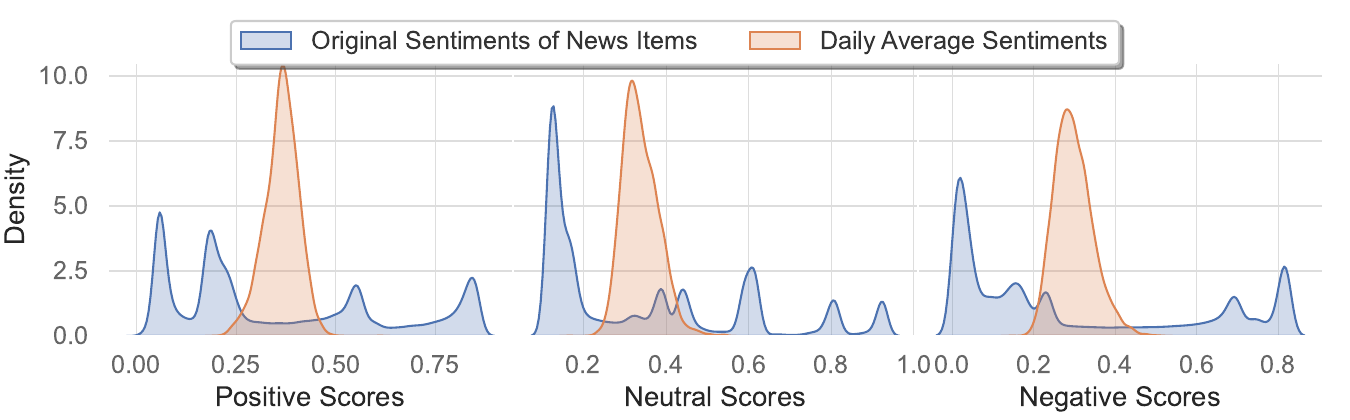}
\label{fig:disribution}
\end{center}
\vspace{-0.5cm}
\end{figure}

To illustrate this critical research gap, we analyzed an extensive financial news dataset spanning from 2003 to 2018, which comprises over 2.7 million news items, with up to 1953 news items per day. Each news item was scored for its potential positive, neutral, and negative impacts on market dynamics. Our investigation reveals a concerning phenomenon we term \textbf{Aggregated Sentiment Homogenization}, which results in weak news sentiment representations. This phenomenon arises when aggregating market sentiments from a large volume of daily news, causing the aggregated sentiment scores to converge towards the average values of the individual sentiment distributions. As a result, the representations of aggregated news sentiments from different days become much more similar (homogenized) compared to the original sentiments of individual news items, thereby smoothing out unique and crucial information. Figure~\ref{fig:disribution} visually presents an overview of this phenomenon, highlighting the notable disparity between the original sentiment scores and the daily average scores. The average score distribution exhibits significantly higher concentration, indicating the loss of information through homogenization. Detailed analysis of this challenge is presented in Section~\ref{sec:issue_identy}

Utilizing homogenized aggregated news sentiment representations for prediction suffers from two limitations. Firstly, given the diverse impact of news on the market, the aggregated scores mask influential news with extreme sentiments and loss crucial predictive information. Secondly, the similarity in average scores across different days homogenizes daily news sentiments, thereby failing to provide effective and distinguishing information for the prediction model. While we use sentiment averaging as an example here, aggregated sentiment homogenization is a broader issue prevalent in many existing aggregation methods. Because it stems from the equal treatment of various news items and static aggregation without considering market changes.

To address these challenges and extract more potential of financial news for prediction, this paper introduces the Market Attention-weighted News Aggregation Network (MANA-Net). MANA-Net innovates by implementing a dynamic market-news attention mechanism that assigns weights to news items based on their correlation with market prices. This approach allows for a nuanced aggregation of news sentiments, emphasizing items with greater predictive relevance. Consequently, MANA-Net can process extensive news of varying volumes and effectively convert it into a unified representation that captures the most predictive sentiment information.  

Furthermore, MANA-Net integrates news aggregation and market prediction into a unified model. This holistic approach enables all model components to dynamically adjust during the training period through gradient back-propagation based on prediction loss. This process allows MANA-Net to optimize the aggregated news sentiment representations by considering price changes and prediction requirements. Therefore, MANA-Net offers an enhanced approach to utilizing market sentiments for market prediction.

In summary, this paper provides three key contributions: 
\setlength{\parskip}{-5pt}
\begin{enumeratesquish}{0em}{1.0em}
\item[(1)] We explore a critical issue inherent in the prevalent equal-weight and static news aggregation methods: aggregated sentiment homogenization. Through analysis of a large financial news dataset, we demonstrate these methods obscure valuable news insights and negatively impact market predictions.

\item[(2)] We introduce MANA-Net, a market prediction model that employs a novel weighted news aggregation approach. MANA-Net mitigates the effects of sentiment homogenization and provides a method to weight news sentiments for market prediction.

\item[(3)] MANA-Net demonstrates superior market prediction performance compared to existing approaches, including recent GPT-based methods. Our model achieves an improvement of 1.1\% in Profit \& Loss and a 0.252 increase in daily Sharpe ratio.
\end{enumeratesquish}
\setlength{\parskip}{0pt}

\section{Related Works}
\label{related_works}
\subsection{Financial news sentiment analysis}
The impact of news on the market has been studied for decades~\cite{wasserfallen1989macroeconomics, veronesi1999stock}. Lexicon-based approaches, utilizing financial keywords dictionaries to extract sentiments, have provided foundational insights and investment guidance~\cite{dewally2003internet, loughran2011liability, feldman2011stock}. However, these methods struggle with complex sentence structures and semantics~\cite{mishev2020evaluation}. This has led to a shift towards machine learning for text extraction~\cite{mirowski2010dynamic, apte2010role,sohangir2018financial}. In addition, deep learning techniques have further enhanced sentiment analysis performance~\cite{severyn2015twitter, zhang2018deep, hu2019transformation}. Techniques like Convolutional Neural Networks (CNNs) and Recurrent Neural Networks (RNNs) have significantly improved sentiment extraction accuracy~\cite{ding2015deep,kraus2017decision}. More recently, language models offer nuanced approaches to analyzing financial news sentiment~\cite{araci2019finbert}. Advanced Large Language Models (LLMs) also develop this topic significantly~\cite{lopez2023can}. Despite these advances in extracting market sentiments from news, the challenge of aggregating sentiments from voluminous news data remains underexplored, particularly in addressing the different impact of individual news items on market predictions.

\subsection{News-based market prediction}
While traditional market prediction models rely on price series data~\cite{gu2020price, xiang2022temporal, koa2023diffusion}, incorporating news data has been proved to provide more valuable insights ~\cite{loughran2011liability}. However, existing studies often employ static and equal-weight aggregation strategies for news sentiments, failing to consider the dynamic nature of markets and the varying impact of individual news items. Early works like sentiment categorization or treating news as a single corpus~\shortcite{li2010information, jin2013forex, li2014news} suffered from limited accuracy due to neglecting the complexities of news language and context. Later advancements like knowledge graphs~\cite{deng2019knowledge} and opinion extraction~\cite{mehta2021harvesting} aimed to extract more information from news data, but still lacked the ability to differentiate news importance. Recent models attempt to combine news and price data, but often resort to simple aggregation techniques like polarity counting~\cite{wang2014crowds, mohan2019stock}, sentiments averaging~\cite{wang2018combining}, or special sentiment indicators designing~\cite{jing2021hybrid, mehta2021harvesting}. This tendency toward simplistic aggregation techniques creates a significant challenge of data homogenization, a problem observed across various domains~\cite{chen2020online, shivaram2022reducing} but underexplored within the context of news-based market prediction. As thousands of market-related news items become readily accessible on a daily basis~\cite{deveikyte2022sentiment}, the inadequacy of current models in effectively harnessing this wealth of information becomes evident, presenting an opportunity for our proposed method, MANA-Net, to enhance market prediction.

\subsection{Attention Mechanism}
The introduction of the attention mechanism in machine translation~\cite{bahdanau2014neural} has revolutionized the ability of models to prioritize crucial information across different domains, including computer vision~\cite{zhu2020deformable} and market selection~\cite{sawhney2021stock}. In essence, the attention mechanism assigns weights to different elements, allowing the model to focus on the most crucial information. This is commonly achieved by comparing a ``query'' vector with a set of ``key'' vectors to compute attention scores. The output is then a weighted sum of corresponding ``value'' vectors. This approach, often referred to as self-attention~\cite{vaswani2017attention}, allows elements within a sequence (e.g., words in a sentence) to attend to each other. The transformer architecture~\cite{vaswani2017attention}, which heavily relies on self-attention, has achieved success in market prediction by extracting insights from price sequences~\cite{liu2019transformer, wang2022stock, zhang2022transformer}. However, these models primarily focus on enhancing knowledge extraction from price data. 
MANA-Net breaks new ground by introducing a novel market-news attention mechanism. Unlike existing applications in market prediction, this mechanism focuses on dynamically estimating weights for each news item. These weights reflect the projected impact of each news item on market volatility. By assigning greater importance to news with a higher anticipated influence on the market, MANA-Net optimizes market prediction through more effective news sentiment utilization. This approach showcases a unique application of attention that extends beyond traditional sequence analysis and allows for the dynamic integration of financial news sentiments.

\section{Preliminaries}
\label{sec:preliminaries}

\subsection{Problem Formulation}
\label{sec:formulation}
The primary goal of this study is to predict the daily market price trends by analyzing daily market prices and news sentiments. For clarity, we introduce the following notations. The trading day is denoted by $d$, and the predicted daily trend, $y_d$, is represented as a binary outcome where 1 signifies a price increase and 0 a not increase. The vector $\boldsymbol{p}_d$ encapsulates fundamental price indicators for the day, with $\boldsymbol{p}_{d}^c$ specifically denoting the close price. The prediction target is formalized as follows:
\begin{equation}
\label{eq:target}
y_d = \mathbb{I} (\boldsymbol{p}_{d+1}^c > \boldsymbol{p}_{d}^c) 
\end{equation}
The set of news sentiments for day $d$ is represented by $\boldsymbol{S}_d = \{\boldsymbol{s}_{d, 1}, \boldsymbol{s}_{d, 2}, ..., \boldsymbol{s}_{d, N_d}\}$, where $\boldsymbol{s}_{d, i}$ indicates the sentiments of the $i$-th news item, and $N_d$ is the total number of news items on day $d$. The function $f(\cdot, \cdot)$ represents the trend prediction model, which integrates market prices and news sentiments of the previous $t$ days ($t\geq1$) as inputs:
\begin{equation}\label{eq:predict}
\begin{aligned}
\hat{y}_d = f((\boldsymbol{p}_d, \boldsymbol{p}_{d-1}, ..., \boldsymbol{p}_{d-t+1}), (\boldsymbol{S}_d, \boldsymbol{S}_{d-1}, ..., \boldsymbol{S}_{d-t+1})) 
\end{aligned}
\end{equation}
Our objective is to effectively learn and optimize the function $f(\cdot, \cdot)$. 

\subsection{Data}
\label{sec:data}
This research utilizes market index datasets: the Standard and Poor's 500 (S\&P 500) Index dataset and the NASDAQ 100 Index dataset, along with the Thomson Reuters News Analytics (TRNA) dataset comprising over 2.7 million news items. Our utilized news dataset is notably larger than those employed in previous studies, with a news volume at least 10 times greater than that of their datasets~\cite{mohan2019stock, jing2021hybrid, lopez2023can}. Moreover, in contrast to their online-collected news data, our dataset is sourced from industry practice, ensuring higher quality and more comprehensive market information.

\noindent \textbf{Market Index Datasets:} The S\&P 500 Index tracks the market performance of 500 leading publicly traded companies in the United States and serves as a critical benchmark for the financial market~\cite{krauss2017deep, fischer2018deep}. Its comprehensive company coverage makes it a pivotal index for evaluating market trends and forecasting. The NASDAQ 100 Index consists of 100 of the largest, most actively traded non-financial companies listed on the Nasdaq Stock Market. It represents the top companies outside of the financial sector and is widely used as a benchmark dataset for stock prediction~\cite{hou2021stock, ding2020hierarchical}. We collected daily S\&P 500 and NASDAQ 100 prices from 2003 to 2018, including six basic features: open price, close price, adjusted close price, high price, low price, and trading volume. Descriptive statistics for the two index datasets are detailed in Table~\ref{table:stats_index}.

\begin{table}[h]
\small
\vspace{-0.2cm}
\caption{Statistics of Market Index Datasets}
\vspace{-0.4cm}
\begin{center}
\begin{tabular}{l p{0.7cm}<{\centering} p{0.9cm}<{\centering} p{0.8cm}<{\centering} p{0.7cm}<{\centering} p{0.9cm}<{\centering} p{0.8cm}<{\centering}}
\toprule
 & \multicolumn{3}{c}{S\&P 500 Index} & \multicolumn{3}{c}{NASDAQ 100 Index} \\
Variable & Mean & Std Dev & Median & Mean & Std Dev & Median \\
\midrule
Open  & 1559.8 & 531.3 & 1367.3 & 2908.9 & 1686.8 & 2197.0 \\
Close & 1560.1 & 531.1 & 1367.6 & 2910.2 & 1687.6 & 2199.1 \\
High  & 1568.3 & 532.0 & 1374.8 & 2931.9 & 1696.9 & 2216.9 \\
Low   & 1550.8 & 530.2 & 1360.8 & 2885.0 & 1675.4 & 2167.3 \\
Adj Close & 1560.1 & 531.1 & 1367.6 & 2910.2 & 1687.6 & 2199.1 \\
Volume & 3.41*$10^9$ & 1.38*$10^9$ & 3.40*$10^9$ & 2.83*$10^5$ & 1.31*$10^5$ & 2.68*$10^5$ \\
\bottomrule
\end{tabular}
\end{center}
\label{table:stats_index}
\vspace{-0.4cm}
\end{table}

\noindent \textbf{News Dataset:} Our utilized TRNA datasets spans from 2003 to 2018. It tracks over 25,000 equities and nearly 40 commodities and energy topics, making it a comprehensive news source for studying financial markets~\cite{mitra2011applications}. All news items are annotated with market sentiment scores by the news provider with an industry's advanced news-analytics. Each news item is assigned three sentiment scores, which indicate the likelihood that the news item will have a positive, neutral, or negative impact on the mentioned instrument. We denote $\boldsymbol{s}_{d, i} = (pos_{d, i}, neu_{d, i}, neg_{d, i})$, where $pos_{d, i}$, $neu_{d, i}$, and $neg_{d, i}$ represent the positive, neutral, and negative scores of the $i$-th news on day $d$. The sum of these three scores for the same news equals one. Statistics of TRNA dataset is illustrated in the next section. 

\section{Sentiment Aggregation Challenges}
\label{sec:issue_identy}
Our investigation into the TRNA dataset has revealed a fundamental challenge inherent to the prevailing equal-weight and static approaches to news sentiment aggregation:  \textbf{Aggregated Sentiment Homogenization}. In essence, assigning equal weight to all news items regardless of their content or potential market impact homogenizes the overall sentiment representations. Furthermore, the static nature of these methods prevents the aggregated representations from considering price data and adapting to different market conditions. Consequently, diverse sentiment knowledge embedded within individual news items are erased, making the aggregated sentiments less informative for detailed market analysis.

\begin{table*}[]
\caption{News sentiment statistics from both TRNA dataset and FinBERT. The 'IQR' refers to the interquartile range. We analyze these statistics at two levels: individual news items and daily aggregated sentiment, including average-based and count-based values. For average aggregation, we calculate the daily average sentiment scores to represent the overall daily sentiments. Count aggregation calculates the proportion of news items belonging to each sentiment category to represent daily sentiments.}
\vspace{-0.4cm}
\begin{center}
\begin{tabular}{l c c c c c c c c c c c c c c}
\toprule
 & \multicolumn{6}{c}{Sentiments from TRNA Dataset} & \quad \quad & \multicolumn{6}{c}{Sentiments from FinBERT} \\ 
Category & Mean & Std Dev & Median & IQR & Skewness & Kurtosis & & Mean & Std Dev & Median & IQR & Skewness & Kurtosis\\
\midrule
\multicolumn{6}{l}{Individual News Items} \\
\hdashline[0.5pt/5pt]
Positive & 0.360 & 0.268 & 0.239 & 0.394 & 0.570 & -1.066 & \multicolumn{1}{l}{} & 0.334 & 0.135 & 0.301 & 0.153 & 3.881 & 1.639 \\
Neutral  & 0.343 & 0.248 & 0.216 & 0.405 & 0.908 & -0.386 & \multicolumn{1}{l}{} & 0.366 & 0.194 & 0.411 & 0.186 & -1.553 & 0.213 \\
Negative & 0.297 & 0.295 & 0.165 & 0.575 & 0.682 & -1.153 & \multicolumn{1}{l}{} & 0.301 & 0.182 & 0.257 & 0.155 & 2.212 & 0.540 \\
\midrule
\multicolumn{11}{l}{Daily Average Values} \\
\hdashline[0.5pt/5pt]
Positive & 0.364 & 0.043 & 0.366 & 0.054 & -0.293 & 0.493 & \multicolumn{1}{l}{} & 0.334 & 0.023 & 0.330 & 0.027 & 2.256 & 18.602 \\
Neutral  & 0.338 & 0.043 & 0.333 & 0.059 & 0.610 & 0.733 & \multicolumn{1}{l}{} & 0.368 & 0.032 & 0.371 & 0.041 & -0.698 & 1.324 \\
Negative & 0.298 & 0.047 & 0.293 & 0.062 & 0.603 & 0.541 & \multicolumn{1}{l}{} & 0.299 & 0.028 & 0.297 & 0.034 & 0.423 & 1.921 \\
\midrule
\multicolumn{11}{l}{Daily Count Ratios} \\
\hdashline[0.5pt/5pt]
Positive & 0.397 & 0.073 & 0.400 & 0.092 & -0.216 & 0.245 & \multicolumn{1}{l}{} & 0.119 & 0.048 & 0.112 & 0.060 & 1.258 & 5.009 \\
Neutral  & 0.273 & 0.076 & 0.264 & 0.101 & 0.628 & 0.674 & \multicolumn{1}{l}{} & 0.659 & 0.072 & 0.664 & 0.098 & -0.417 & 1.204 \\
Negative & 0.330 & 0.071 & 0.324 & 0.094 & 0.526 & 0.433 & \multicolumn{1}{l}{} & 0.222 & 0.063 & 0.219 & 0.084 & 0.331 & 2.024 \\
\bottomrule
\end{tabular}
\end{center}
\label{table:stats_news}
\vspace{-0.6cm}
\end{table*}

There are various widely used news sentiment aggregation representations such as Count Features, Sentiment Factors and Average Features~\cite{wang2018combining, jing2021hybrid, lopez2023can}, which will be introduced in Section~\ref{sec:baselines}. Existing aggregated representations are primarily based on two core indicators, the average sentiment scores and the news polarity counts. Both of them can lead to the homogenization issue, affecting all related aggregation methods. To demonstrate the prevalence of this challenge beyond our specific news sentiments from TRNA dataset, we collected additional sentiment scores using FinBERT~\cite{araci2019finbert}, a widely recognized area-specific language model proficient in financial sentiment analysis. We generated FinBERT scores for our news. Table~\ref{table:stats_news} presents statistics of two sentiment sets for analysis.

Statistics of individual news items reveal that sentiment scores are widely scattered, rendering overall statistics less representative. Considering the 0-1 scoring range, the high standard deviations (around 0.25 and 0.15 for both datasets) indicate significant dispersion. Furthermore, the substantial gaps between mean and median values, along with large interquartile ranges (IQR), support a relatively even distribution across a broad range. Skewness and kurtosis provide further insights into the distribution shapes.  TRNA sentiment scores exhibit skewness and significant platykurticity. FinBERT scores, while less platykurtic, still demonstrate a strong skew. These characteristics – high dispersion, wide ranges, and skewed distributions –  pose challenges for existing aggregation methods that rely heavily on average scores or counts. 

\begin{figure}[b]
\centering
\vspace{-0.6cm}
\begin{center}
   \captionof{figure}{The boxplot of positive sentiment scores for individual news items across 30 randomly sampled days. Green triangles represent the daily average scores, while red lines indicate the median values. The blue rectangle highlights the range of averaged sentiments.}
   \includegraphics[width=0.98\linewidth]{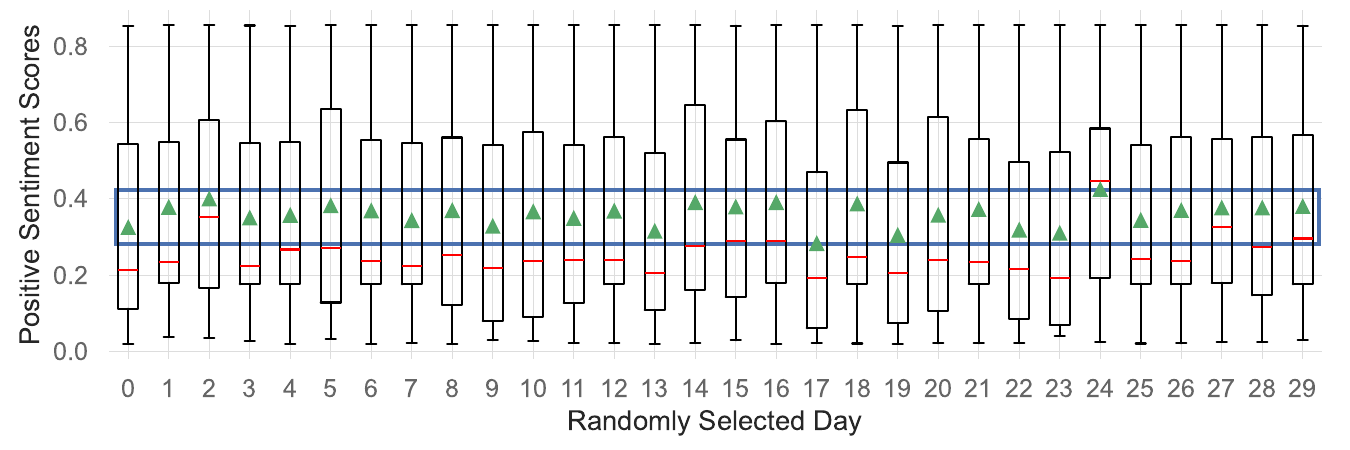}
\label{fig:boxplot}
\end{center}
\vspace{-0.0cm}
\end{figure}

Our analysis of aggregated sentiment scores from both TRNA and FinBERT highlights a concerning trend: homogenization. After aggregation, the standard deviation values for daily sentiment on both datasets drop below 0.1, with most values even less than 0.05. This represents a decrease of more than 80\% compared to the standard deviation of individual news item scores. Similarly, the difference between mean and median values, along with the interquartile range (IQR), all shrink significantly. These observations suggest that the aggregated sentiment scores become highly concentrated. Kurtosis values further support this finding. Compared to the original sentiment scores, the kurtosis values of the aggregated sentiment scores increase significantly. This indicates that the distributions become peaked and more concentrated around the average value. These observations suggest that the aggregated sentiment scores become highly concentrated, losing the variability present in the original data and providing less distinguished representations for the following prediction models. 

To complement our statistical analysis, we have included the distribution figure in Section~\ref{sec:introduction}. In addition, we provide a boxplot (Figure~\ref{fig:boxplot}) to offer a visual representation of our findings. We randomly selected 30 days from the dataset and drew a boxplot of the positive news scores for these days. The figure clearly illustrates the homogenization effect of aggregation. As observed, all daily average scores (represented by green triangles) fall within a narrow value range (highlighted by the blue rectangle). In contrast, the boxplot for individual news item scores encompasses a much wider range, visually demonstrating the greater variability in sentiment before aggregation.

In summary, our investigation exposes a fundamental challenge in current sentiment aggregation methods: Aggregated Sentiment Homogenization. This issue obscures the valuable diversity of sentiments within individual news articles, leading to the generation of uninformative aggregated sentiment representations.

\section{Methodology}
\label{sec:method}

\begin{figure*}[h]
\centering
\begin{center}
   \captionof{figure}{An overview of MANA-Net. The market prices and news sentiments data are as defined in Section~\ref{sec:formulation}. Blue arrows show the feed-forward process and grey arrows show the back-propagation process. }
   \vspace{-0.0cm}
   \includegraphics[width=0.9\linewidth]{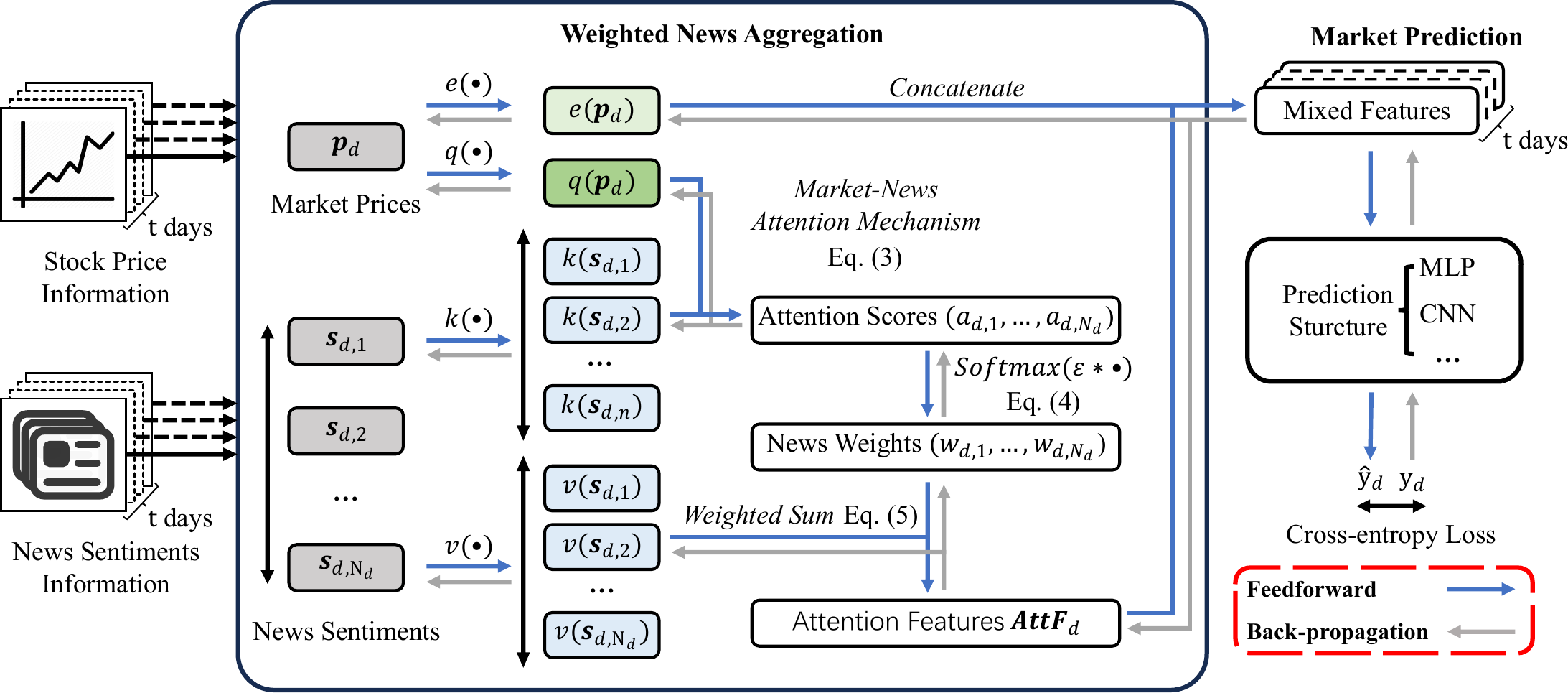}
\label{fig:model}
\end{center}
\vspace{-0.5cm}
\end{figure*}
MANA-Net tackles the challenge of aggregated sentiment homogenization by incorporating a trainable news aggregation process within the prediction model itself. It takes both daily market prices and sentiments of individual news item as inputs. A core component of MANA-Net is a market-news attention mechanism. This mechanism dynamically evaluates the relevance between each news item and the daily market prices, assigning news weights that determine the impact of each news item on the prediction outputs. These importance scores are then used to  weight individual sentiment scores during aggregation.  By integrating news aggregation and market prediction into a unified trainable model, MANA-Net ensures that the aggregation process adapts to the prediction objective, mitigating sentiment homogenization and generating a more informative representation for superior market prediction performance. Figure~\ref{fig:model} presents an overview of MANA-Net.

\subsection{Weighted News Aggregation}
MANA-Net's weighted news aggregation component is specifically designed to address the issue of aggregated sentiment homogenization. It meets three key requirements for effective news sentiment aggregation: (i) quantifying the impact of news on the market based on price changes, (ii) generating more effective aggregated representations incorporating the market impact of news, and (iii) accommodating fluctuating news volumes.

Firstly, to quantify the impact of news, we design a market-news attention mechanism. Traditional self-attention mechanisms focus on a sequence's internal information to weight its content~\cite{vaswani2017attention}. In financial areas, such mechanisms are used for price series analysis by attending to prices at different points~\cite{ding2020hierarchical}. MANA-Net innovates by applying a split-attention approach to distinct input information. Specifically, we prepare functions for query, key, and value vectors, denoted as $q(\cdot)$, $k(\cdot)$, and $v(\cdot)$, which encodes inputs into the corresponding features. Unlike traditional methods, these functions are applied to different data sources. As shown in Figure~\ref{fig:model}, $q(\cdot)$ operates on market price data $\boldsymbol{p}_d$, while $k(\cdot)$ is used on news sentiment data $(\boldsymbol{s}_{d, 1}, \boldsymbol{s}_{d, 2}, ..., \boldsymbol{s}_{d, N_d})$. We then employ scaled dot-product attention~\cite{vaswani2017attention} to calculate attention scores $a_{d, i}$ as:
\begin{equation}
\label{eq:att1}
\begin{aligned}
a_{d, i} = \frac{1}{\sqrt{d_k}}q(\boldsymbol{p}_d)^T \cdot k(\boldsymbol{s}_{d, i}) 
\end{aligned}
\end{equation}
where $d_k$ denotes the key vector dimension, using for scaling the products of $q(\boldsymbol{p}_d)$ and $k(\boldsymbol{s}_{d, i})$ to prevent exploding gradients during training. This design leverages the core principle of attention mechanisms: measuring relevance between query and key vectors to weight key-corresponding data. In MANA-Net, this separation allows news information to be evaluated based on market prices. The resulting attention scores fulfill the first goal by quantifying news impact based on price changes.

Secondly, we design a difference enlargement normalization to translate attention scores into news weights. Large daily news volumes can lead to less differentiated values when directly applying the softmax function to attention scores. To address this and accelerate training convergence, MANA-Net introduces a difference enlargement factor $\epsilon (\epsilon \geq 1)$ within the softmax function. We normalize attention scores to news aggregation weights $\boldsymbol{W}_d = (w_{d,1}, w_{d,2}, ..., w_{d,N_d})$ as :
\begin{equation}
\label{eq:att2}
\boldsymbol{W}_d = Softmax(\epsilon \cdot (a_{d,1}, a_{d,2}, ..., a_{d,N_d}))
\end{equation}
This factor emphasizes the relative importance of news items by enhancing the distinction between news weights while preserving their rankings. The factor $\epsilon$ is tuned as a hyperparameter during training. The resulting weights $\boldsymbol{W}_d$ determine the contribution of each news item to the aggregated representation. They are used in a weighted sum over the value vectors $v(\boldsymbol{s}_{d, i})$, generating a unified vector called \textbf{Attention Features (AttF)}.
\begin{equation}
\label{eq:att3}
\boldsymbol{AttF}_d = \Sigma_i^{N_d} w_{d,i} v(\boldsymbol{s}_{d, i})
\end{equation}
This $\boldsymbol{AttF}_d$ vector captures the sentiment information from all news on day $d$, while weighting their contributions based on market impact. To this end, we achieve our second goal of generating more effective aggregated representations for market prediction.

Thirdly, upon reviewing the process of MANA-Net's weighted news aggregation, it becomes evident that there are no constraints on news volume. Equations (\ref{eq:att1}), (\ref{eq:att2}), and (\ref{eq:att3}) operate with any value of $N_d$. The market-news attention mechanism and the weighted news aggregation operations are adaptable to varying news volumes. There is no need for time-consuming and potentially biased process of news selection or filtering, common in approaches that struggle with fluctuating news volumes~\cite{huang2010realization, li2010information}. Consequently, the third goal of accommodating fluctuating news volume is achieved.

In summary, MANA-Net's market-news attention mechanism assesses the importance of news to the market, and its weighted news aggregation emphasizes highly influential news. This process effectively addresses the challenges of sentiment homogenization by ensuring that critical news items with extreme positive or negative sentiment scores are not simply smoothed over. Capturing these extremes allows the model to gain a nuanced understanding of market sentiment, crucial for effective market prediction.

\subsection{Market Prediction and Training}
\label{sec:method_pre}
MANA-Net further addresses the issue of sentiment homogenization by an integrated market prediction model that incorporates our news aggregation component. Existing works typically treat aggregated news sentiments as a static input~\cite{mehta2021harvesting,li2022novel, lopez2023can}. Therefore, they are limited by unchangeable news representations, which has been homogenized due to the equal-weight aggregation. Unlike these methods, MANA-Net creates a trainable news aggregation process directly connected to the prediction model. This holistic approach treats the aggregation and prediction components as a unified system, taking both prices and original news sentiments as inputs to output market predictions.

Specifically, once the aggregated news sentiment vector, $\boldsymbol{AttF}_d$, is obtained, it is concatenated with $e(\boldsymbol{p}_d)$, a distinct price feature vector designed for prediction. The function $e(\cdot)$ here is an encoding function different from $q(\cdot)$, allowing for potentially different encoding strategies for different data utilization. The concatenated vector $\boldsymbol{m}_d = [e(\boldsymbol{p}_d), \boldsymbol{AttF}_d]$ serves as the input for the prediction model. If the prediction leverages multi-day data (previous $t$ days, where $(t>1)$), we create a sequence of features $(\boldsymbol{m}_d, \boldsymbol{m}_{d-1}, ..., \boldsymbol{m}_{d-t+1})$ for each day within the chosen range. The prediction model then utilizes a cross-entropy loss function, commonly used for classification tasks, to quantify prediction error for training.

This seamless integration of news aggregation with prediction facilitates the optimization of both components in alignment with prediction objectives. Unlike static aggregation methods where backpropagation terminates at the aggregated representations, MANA-Net allows gradients to flow through the entire process. This enables the model to learn from prediction errors and adjust the weights throughout the news aggregation pipeline. The backpropagation process can be illustrated by the following equation :
\begin{equation}
\label{eq:gradients}
\frac{\partial y}{\partial \boldsymbol{s}_{d, i}} = \frac{\partial y}{\partial \boldsymbol{ATTF_d}} \cdot \frac{\partial \boldsymbol{ATTF}_d}{\partial \boldsymbol{s}_{d, i}}
\end{equation}
This equation delineates how the error signal (represented by the partial derivative of the loss function with respect to the prediction $y$) is propagated backwards. The dynamic adjustment allows the model to modify the news aggregation process based on prediction errors, channeling gradients from the aggregated representations $\boldsymbol{AttF}_d$ back to individual news sentiments $\boldsymbol{s}_{d, i}$. Such a mechanism ensures that the news feature vector continuously evolves to encapsulate the most relevant information for market prediction. Figure~\ref{fig:model} visually depicts the training process workflow, highlighting the forward pass (blue arrows) and backward pass (grey arrows) of data and gradients. This integrated training approach ensures that all model components, including the news aggregation mechanism, are optimized for better market prediction.

MANA-Net is adaptable and can be used with various neural network architectures for prediction.  We evaluate its performance using five popular structures: shallow network (SN), Multi-Layer Perceptron (MLP)~\cite{cybenko1989approximation}, Convolutional Neural Network (CNN)~\cite{lecun1989backpropagation}, Long Short-Term Memory (LSTM) model, and Transformer (Trans) structure~\cite{vaswani2017attention}. These architectures, widely acknowledged in market prediction literature~\cite{jing2021hybrid,jiang2021applications}, are evaluated through time series cross-validation to identify the optimal structure for MANA-Net. For SN, MLP, and CNN models, the mixed features are concatenated into a single vector before being fed into the prediction model. LSTM and Transformer structures, designed for sequential data, take the vectors $(\boldsymbol{m}_{d-t+1}, \boldsymbol{m}_{d-t+2}, ..., \boldsymbol{m}_d)$ as input in sequential order. A detailed analysis of the prediction structures will be presented in Section~\ref{sec:preidiction_model}.

\section{Experiment Settings}
We conducted experiments using both 2080Ti and A100 GPUs. Training a model requires 28 hours on a 2080Ti GPU or 6 hours on an A100 GPU. Trained models run for over a month in our tests, requiring only 9.45 seconds for a single prediction. Given the common usage of A100 and superior GPUs in the industry, our computing requirements are practical for real-world applications.

\subsection{Evaluation Metrics}
We evaluate our model based on three metrics: \textbf{Accuracy}, \textbf{Profit \& Loss (PnL)} and \textbf{Sharpe Ratio (SR)}. They are widely used in market prediction literature~\cite{ye2020reinforcement, yuemei2021predicting}. Among these three metrics, SR is considered the most important one as it incorporates both returns and risk, which are crucial factors in market investment.

\noindent \textbf{PnL} quantifies the cumulative profit or loss experienced by a portfolio during a designated time frame. We calculate the PnL for each trading day and sum up the daily PnL values across the entire test sets. The PnL for all forecasts spanning $D$ days can be expressed as:
\begin{equation}
\label{eq:pnl}
\textit{PnL} = \sum^D_{d=1} \textit{flag}_d \cdot \frac{\boldsymbol{p}^c_{d+1} - \boldsymbol{p}^c_{d}}{\boldsymbol{p}^c_{d}},
\end{equation}
where $\textit{flag}_d=1$ if $\hat{y}_d = y_d$ and $\textit{flag}_d=-1$ otherwise.

\noindent \textbf{Sharpe Ratio} measures the investment performance in relation to a risk-free asset. To compute the daily Sharpe Ratio within the test sets, we employ the following formula:
\begin{equation}
\label{eq:sharpe_ratio2}
\textit{SR} = \left(\displaystyle\frac{1}{n}\displaystyle\sum_{d=1}^D \displaystyle flag_d\frac{\boldsymbol{p}^c_{d+1} - \boldsymbol{p}^c_{d}}{\boldsymbol{p}^c_{d}} - R_f\right)  \Bigg/ {\sigma(R)},
\end{equation}
where $R_f$ represents the return rate of an investment with zero risks, meaning that it is the return that investors could expect for taking no risk, such as a Treasury bond investment. And $\sigma(R)$ denotes the standard deviation of the excess return of the asset. In our settings, we use a value of 0.02 for $R_f$, which corresponds to the average US Treasury rate during our data collection period. Furthermore, we also explored different risk-free rate values between 0.00 and 0.05.  MANA-Net exhibited robustness to this parameter, consistently achieving superior performance compared to baseline methods.

\subsection{Baseline Methods}
\label{sec:baselines}
To assess MANA-Net's news aggregation capabilities, we compare it with five popular news aggregation methods:

\begin{enumeratesquish}{0em}{-0.5em}
\item[(1)] \textbf{Count Features (CF)}: This common method calculates ratios of news categories' counts to represent overall sentiment (e.g., positive news articles / total news articles)~\cite{wang2018combining, lopez2023can}.

\item[(2)] \textbf{Sentiment Factor (SenF)}: Proposed by Jing et al.~\shortcite{jing2021hybrid}, SenF is a sentiment factor for LSTM model to predict the market trends. SenF is calculated as: $SenF_d = (num_d^{pos} - num_d^{neg}) / num_d^{pos\&neg}$,
where $num_d^{*}$ is the volume of news belonging to class * on day $d$.

\item[(3)] \textbf{Sum Features (SumF)}: This method simply sums the daily news sentiment scores to create a single sentiment vectors~\cite{mohan2019stock}.

\item[(4)] \textbf{Average Features (AF)}: Similar to SumF, AF aggregates sentiment by averaging daily individual news sentiment scores~\cite{wang2018combining}.

\item[(5)] \textbf{Frequency Aggregation Features (FAF):} Huynh et al.~\shortcite{huynh2017stock} proposed FAF, which assigns weights based on news appearance frequency, considering the potential impact of news repetition.
\end{enumeratesquish}

\noindent When comparing these methods to MANA-Net, we implemented these approaches by substituting our news aggregation model and utilizing their news representations instead of our \textbf{AttF}. The inputs and prediction structures remain the same to ensure a fair comparison. Additionally, since these methods' original prediction structures are included in our prediction structure candidates, they also serve as market prediction baselines compared to MANA-Net.

In addition to news aggregation, we evaluate MANA-Net's overall market prediction performance against three advanced methods:

\begin{enumeratesquish}{0em}{-0.5em}
\item[(6)] \textbf{Ensemble Deep Learning Model}~\cite{li2022novel}: This model integrates sentiment analysis with a two-level ensemble architecture, known for its effectiveness in capturing complex time series patterns and combining textual and numerical data.

\item[(7)] \textbf{Frozen Pretrained Transformer (FPT)}~\cite{zhou2023one}: FPT leverages a pre-trained Transformer model, demonstrating state-of-the-art performance in various time series tasks. 

\item[(8)]\textbf{GPT-based Methods}~\cite{lopez2023can}:This recent work explores the potential of LLMs like GPT-3.5 and GPT-4 for market prediction using news headlines. It highlights the growing interest in leveraging these powerful language models for financial forecasting.

\end{enumeratesquish}

\noindent These market prediction methods covers the advanced time-series models, ensemble models, and large language models. Therefore, they represent the state-of-the-art methods of market prediction. 

\subsection{Training and Validation}
We assess the performance of all methods using data from the preceding $t$ days. Our experimental design involves varying values for $t$, including 1, 3, 5, 10, and 20. To ensure a rigorous and comprehensive assessment, we adopt a time series cross-validation approach to determine the optimal hyper-parameters. The dataset is partitioned into 10 sliding windows, each encompassing 500 consecutive days of data, arranged chronologically. The initial window starts from the first available day, with subsequent windows commencing 391 days later than their predecessor. Within each window, we allocate the data into training, validation, and testing sets following an 8:1:1 ratio, respectively. The presented results are averages from the test sets across all 10 windows, ensuring stability and reliability across different temporal segments of the dataset.

\section{Results and Discussions}
In our experiments, using sentiments derived from TRNA datasets generally yields better market prediction results compared to sentiments obtained from other methods like FinBERT. Therefore, our discussions are primarily based on results with TRNA sentiments.

\subsection{Prediction model Selection}
\label{sec:preidiction_model}
To determine the most effective prediction structure for MANA-Net, we maintain the weighted news aggregation model constant while experimenting with candidate prediction structures. The results are summarized in Table~\ref{tab:prediction_model}. Notably, the MLP emerges as the top-performing structure across most tested settings in terms of SR and PnL, establishing itself as the most suitable choice. Despite their renown for handling time-series data, both the LSTM model and the Transformer structure do not outperform the other models.

\begin{table}[h]
\vspace{-0.2cm}
\renewcommand\arraystretch{1}
\captionof{table}{MANA-Net's prediction results with different prediction structures. Refer to Section~\ref{sec:method_pre} for model abbreviations.}
\vspace{-0.4cm}
\begin{center}
\begin{tabular}[b]{p{1.2cm}<{\centering} c c c c c c}
\toprule
\multirow{2}{*}{\shortstack{Time\\Length}}  & \multirow{2}{*}{Metrics}
& \multicolumn{5}{c}{Model Architectures} \\
\cmidrule(r){3-7}
&\quad
& SN & MLP & CNN & LSTM & Trans \\
\cmidrule(r){1-7}
\multirow{3}{*}{1-day} 
    & Acc & 0.535 & \textbf{0.537} & 0.518 & 0.529 & 0.534 \\
    & PnL & 0.022 & \textbf{0.031} & 0.017 & 0.007 & 0.020 \\
    & SR  & 0.499 & \textbf{1.110} & 0.576 & 0.382 & 0.599 \\
\cmidrule(r){1-7}
\multirow{3}{*}{3-day} 
    & Acc & 0.535 & 0.541 & \textbf{0.543} & 0.529 & 0.507 \\
    & PnL & 0.004 & 0.037 & \textbf{0.053} & 0.011 & 0.032 \\
    & SR  & 0.080 & 0.943 & \textbf{1.565} & 0.557 & 0.692 \\
\cmidrule(r){1-7}
\multirow{3}{*}{5-day} 
    & Acc & 0.525 & \textbf{0.542} & 0.507 & 0.530 & 0.511 \\
    & PnL & 0.024 & \textbf{0.041} & 0.005 & 0.008 & 0.031 \\
    & SR  & 0.180 & \textbf{1.331} & -0.028 & 0.416 & 0.831 \\
\cmidrule(r){1-7}
\multirow{3}{*}{10-day} 
    & Acc & \textbf{0.563} & 0.552 & 0.516 & 0.530 & 0.503 \\
    & PnL & 0.045 & \textbf{0.054} & 0.004 & 0.008 & 0.029 \\
    & SR  & 1.317 & \textbf{1.624} & -0.192 & 0.416 & 0.563 \\
\cmidrule(r){1-7}
\multirow{3}{*}{20-day} 
    & Acc & \textbf{0.550} & 0.528 & 0.526 & 0.503 & 0.509 \\
    & PnL & 0.023 & \textbf{0.045} & 0.029 & 0.029 & 0.033 \\
    & SR  & 0.834 & \textbf{1.533} & 0.608 & 0.597 & 0.747 \\
\bottomrule
\end{tabular}
\end{center}
\label{tab:prediction_model}
\vspace{-0.4cm}
\end{table}

\begin{table*}[]
\renewcommand\arraystretch{1}
\captionof{table}{Market prediction results of prevalent aggregation methods and advanced prediction methods. Abbreviations for these methods are detailed in Section~\ref{sec:baselines}. The ``Price Only'' lines serve as the control group, relying solely on prices for predictions. Results for the GPT-4 on the NASDAQ 100 dataset are exclusive due to its less competitive results on S\&P 500 dataset and its limited availability compared to GPT-3.5. Underlined values highlight the highest PnL and SR achieved across all settings.}
\vspace{-0.4cm}
\begin{center}
\begin{tabular}[b]{l 
c@{\hspace{4pt}} c@{\hspace{4pt}} c@{\hspace{15pt}} 
c@{\hspace{4pt}} c@{\hspace{4pt}} c@{\hspace{15pt}}  
c@{\hspace{4pt}} c@{\hspace{4pt}} c@{\hspace{15pt}}  
c@{\hspace{4pt}} c@{\hspace{4pt}} c@{\hspace{15pt}} 
c@{\hspace{4pt}} c@{\hspace{4pt}} c }
\toprule
\multirow{2}{*}{\shortstack{Method}} 
& & 1-day & & & 3-day & & & 5-day & & & 10-day & & & 20-day & \\
& Acc & PnL & SR & Acc & PnL & SR & Acc & PnL & SR & Acc & PnL & SR & Acc & PnL & SR \\
\cmidrule(r){1-16}
\multicolumn{16}{l}{Results of S\&P 500 Market} \\
\cmidrule(r){1-16}
Price Only & 0.535 & 0.026 & 0.870 & 0.512 & 0.030 & 0.886 & 0.525 & 0.024 & 0.592 & 0.531 & 0.029 & 0.820 & 0.546 & 0.026 & 0.686 \\
CF       & 0.498 & 0.024 & 0.619 & 0.533 & 0.008 & 0.309 & 0.536 & 0.015 & -0.051 & 0.530 & 0.021 & 0.621 & 0.526 & 0.024 & 0.915 \\
SenF     & \textbf{0.539} & 0.021 & 0.767 & 0.496 & 0.010 & -0.164 & \textbf{0.554} & 0.030 & 1.083 & 0.525 & 0.038 & 1.013 & \textbf{0.554} & 0.045 & 1.085 \\
SumF     & 0.512 & -0.022 & -0.637 & 0.514 & 0.019 & 0.276 & 0.503 & -0.019 & -0.409 & 0.501 & 0.024 & 0.814 & 0.513 & 0.036 & 0.853 \\
AF       & 0.535 & 0.017 & 0.515 & \textbf{0.541} & \textbf{0.037} & 0.943 & 0.513 & -0.004 & -0.098 & 0.509 & 0.026 & 0.732 & 0.515 & 0.023 & 0.712 \\
FAF      & 0.518 & 0.018 & 0.389 & 0.534 & 0.031 & 0.845 & 0.515 & -0.016 & -0.716 & 0.511 & 0.012 & 0.339 & 0.503 & 0.029 & 0.597 \\
\textbf{MANA-Net} & 0.537 & 0.031 & \textbf{1.110} & 0.518 & 0.029 & 0.978 & 0.542 & 0.041 & \textbf{1.331} & \textbf{0.552} & \underline{\textbf{0.056}} & \underline{\textbf{1.624}} & 0.528 & \textbf{0.045} & \textbf{1.533} \\
Ensemble & 0.522 & 0.018 & 0.886 & 0.514 & 0.031 & 0.672 & 0.510 & 0.018 & 0.824 & 0.534 & 0.031 & 1.343 & 0.545 & 0.028 & 1.028 \\
FPT      & 0.508 & 0.002 & 0.323 & 0.524 & 0.033 & 0.917 & 0.520 & \textbf{0.042} & 1.169 & 0.528 & 0.025 & 0.969 & 0.547 & 0.035 & 1.273 \\
GPT-3.5  & 0.522 & 0.022 & 0.914 & 0.530 & 0.030 & \textbf{1.133} & 0.528 & 0.022 & 0.946 & 0.522 & 0.014 & 0.765 & 0.524 & 0.014 & 0.769 \\
GPT-4    & 0.504 & \textbf{0.032} & 0.696 & 0.520 & 0.008 & 0.194 & 0.516 & 0.008 & -0.104 & 0.484 & 0.003 & -0.244 & 0.494 & 0.005 & -0.109 \\
\cmidrule(r){1-16}
\multicolumn{16}{l}{Results of NASDAQ 100 Market} \\
\cmidrule(r){1-16}
Price Only & 0.512 & -0.021 & -0.834 & 0.498 & -0.057 & -1.537 & 0.517 & -0.010 & -0.476 & 0.557 & 0.024 & 0.950 & 0.517 & -0.008 & 0.016 \\
CF & 0.548 & 0.010 & 0.711 & 0.482 & -0.045 & -1.227 & 0.527 & -0.003 & 0.136 & 0.533 & -0.005 & 0.231 & 0.507 & -0.023 & -0.475 \\
SenF & 0.528 & \textbf{0.024} & 0.150 & 0.510 & -0.020 & -0.257 & \textbf{0.569} & \textbf{0.023} & \textbf{0.937} & 0.505 & -0.035 & -0.594 & 0.517 & -0.016 & -0.381 \\
SumF & 0.536 & 0.010 & 0.570 & 0.514 & -0.019 & -0.217 & 0.545 & 0.006 & 0.588 & 0.527 & 0.009 & 0.457 & 0.540 & 0.014 & 0.883 \\
AF & 0.504 & -0.035 & -0.984 & 0.536 & -0.013 & 0.059 & 0.521 & 0.008 & 0.306 & 0.513 & -0.026 & -0.516 & 0.491 & -0.024 & -0.884 \\
FAF & \textbf{0.552} & -0.002 & 0.401 & 0.536 & -0.034 & -0.425 & 0.539 & -0.016 & -0.101 & 0.525 & -0.016 & -0.120 & 0.527 & -0.006 & 0.059 \\
\textbf{MANA-Net} & 0.528 & 0.004 & 0.592 & \textbf{0.550} & \textbf{0.023} & \textbf{0.891} & 0.549 & 0.020 & 0.901 & \textbf{0.561} & \textbf{0.031} & \textbf{1.017} & 0.538 & \underline{\textbf{0.041}} & \underline{\textbf{1.347}} \\
Ensemble & 0.540 & -0.006 & 0.107 & 0.534 & 0.007 & 0.628 & 0.546 & 0.010 & 0.627 & 0.532 & 0.020 & 0.985 & \textbf{0.542} & 0.024 & 1.095 \\
FPT & 0.536 & 0.009 & 0.523 & 0.537 & 0.003 & 0.457 & 0.549 & 0.009 & 0.766 & 0.540 & 0.017 & 0.837 & 0.506 & 0.003 & 0.322 \\
GPT-3.5 & 0.538 & 0.010 & \textbf{0.795} & 0.522 & 0.014 & 0.882 & 0.526 & 0.015 & 0.692 & 0.530 & 0.017 & 0.896 & 0.522 & 0.009 & 0.639 \\
\bottomrule
\end{tabular}
\end{center}
\label{tab:main_results}
\vspace{-0.5cm}
\end{table*}

This outcome can be attributed to the complexity of the model. The LSTM model has been criticized for its intricate structure, characterized by multiple gates and parameters lacking immediate interpretability or clear purpose~\cite{jozefowicz2015empirical}. This complexity poses challenges to the training process. In the context of MANA-Net, which already incorporates a sophisticated news aggregation structure, a highly complex prediction network could potentially hinder overall performance. Our validation results support this notion, as the most effective LSTM configuration involved using a single layer and one-directional model, the simplest form. Similarly, the inherent complexity of Transformer does not align well with MANA-Net. The best validation results obtained from a transformer with 2 attention layers, whereas general transformers typically employ 6/12 layers. Thus, we require a prediction structure with an optimal level of complexity that complements our weighted news aggregation.

Additionally, while LSTM and Transformer are renowned for their ability to capture sequential information, other structures can also leverage temporal dependencies. Several studies have successfully applied MLP models to time series prediction tasks, showcasing their capacity to use time-dependent message from input data~\cite{maia2011holt, farnoosh2021deep, orimoloye2020comparing}. Moreover, when provided with high-quality features, MLP models are frequently employed as subsequent layers due to their simplicity~\cite{orimoloye2020comparing, hu2020hrn}. Given the concatenated features, the MLP model can inherently learn the significance of time order if it is critical for the prediction. Therefore, the MLP model, with its appropriate level of complexity, appears to be the most suitable prediction structure for our MANA-Net framework.

\subsection{Prediction Results and Analysis}

We assess MANA-Net against various news aggregation methods and contemporary market prediction models, illustrating its effectiveness in harnessing news for market predictions. Comparative results, shown in Table~\ref{tab:main_results}, use the 'Price Only' method as a benchmark to demonstrate outcomes without integrating news data. Abbreviations for baseline methods are consistent with those introduced in Section~\ref{sec:baselines}. Results for the GPT-4 on the NASDAQ 100 dataset are exclusive due to its less competitive results on S\&P 500 dataset and its limited availability compared to GPT-3.5.

\vspace{-0.2cm}
\subsubsection{Comparison with Aggregation Methods}
Our evaluation of MANA-Net against existing news aggregation methods (presented in the upper side of MANA-Net in Table~\ref{tab:main_results}) reveals three key findings that highlight MANA-Net's more effective news utilization. (i) None of the baseline methods consistently outperform the ``Price Only'' strategy, indicating that inappropriate news aggregation methods may not only fail to improve but could even negatively affect the prediction. (ii) No single baseline method demonstrates superior performance across all scenarios, highlighting a general lack of robustness.
(iii) MANA-Net demonstrates consistent improvement, outperforming the control group in all settings and surpassing baselines. It achieves the best performance in most scenarios and achieves the highest PnL and SR across all settings on both datasets. This consistent improvement highlights MANA-Net's effectiveness in utilizing news for prediction. Notably, at the optimal settings for each methods, MANA-Net outperforms baselines by at least 1.1\% in PnL and 0.539 in SR for the S\&P 500 market, and 1.6\% in PnL and 0.397 in SR for the NASDAQ 100 market.

The comparison with existing news aggregation methods underscores the effectiveness of MANA-Net's weighted news aggregation. The key difference lies in how news aggregated representations are obtained. Most existing methods, except for ``FA'', treat news articles equally and use static representations for prediction, making them susceptible to sentiment homogenization issue. The ``FAF'' method partially addresses this issue by considering news frequency as a proxy for news importance, but it still relies on static representations. MANA-Net's superior performance over these methods highlights its more effective news aggregation in overcoming sentiment homogenization (see Section~\ref{sec:analysis_weights} for detailed analysis).

\begin{table*}[]
\renewcommand\arraystretch{1}
\captionof{table}{News Cases with Weights Exceeding 0.98. This table lists specific news items that received weights greater than 0.98 after normalizing daily weights to a 0-1 scale, along with their sentiment scores. The sentiment scores are presented in the format of (positive scores, neutral score, negative score).}
\vspace{-0.4cm}
\begin{center}
\begin{tabular}[b]{c p{9.6cm} c c}
\toprule
Date & News Title & News Sentiments & Daily Avg. Sentiments\\
03/10/2018 & Reuters Insider - Trading at Noon: A look at what is powering Dow Industrials to another record & (0.851, 0.118, 0.031) & (0.373, 0.302, 0.325) \\
05/10/2018 & Macron to campaign for tougher anti-monopoly rules in EU elections & (0.057, 0.125, 0.819) & (0.393, 0.325, 0.282) \\
12/10/2018 & Saudis face business backlash over missing journalist
 & (0.056, 0.126, 0.818) & (0.347, 0.320, 0.332) \\
16/10/2018 & Pompeo meets Saudi king on Khashoggi case, Turks study "toxic materials" & (0.058, 0.134, 0.808) & (0.375, 0.309, 0.317) \\
\bottomrule
\end{tabular}
\end{center}
\label{tab:case_study}
\vspace{-0.5cm}
\end{table*}

\vspace{-0.2cm}
\subsubsection{Comparison with Advanced Market Prediction Methods}

MANA-Net's strength in leveraging news knowledge positions it as a frontrunner among contemporary market prediction models. It surpasses Ensemble and FPT methods, which also exhibit commendable and stable performance, particularly when incorporating multi-day data. MANA-Net achieves a PnL improvement of 1.4\% and an SR increase of 0.281 on the S\&P 500 market and a 1.7\% PnL gain and a 0.252 SR boost on the NASDAQ 100 market compared to these models. The advanced GPT-based methods underperform in our settings, primarily due to the significantly larger size of our news dataset compared to those used in their original development. 
GPT-based methods, which rely on polarity counts for news aggregation, suffer from the sentiment homogenization issues. For instance, GPT-3.5 categorizes a substantial portion (55.38\%) of news articles in the S\&P 500 dataset as neutral, with 29.25\% classified as positive and 15.36\% as negative, leading to highly homogenized aggregated news sentiments that are predominantly neutral. Similarly, GPT-4 classifies 66.78\% of news as neutral, further exacerbating the homogenization issue. The NASDAQ 100 dataset displays a similar trend as well. These observations underscore the necessity for a more nuanced approach to news aggregation. By effectively parsing and weighting news sentiments, MANA-Net proves to be a superior solution for news-driven market prediction.

\vspace{-0.1cm}
\subsection{Analysis of News Weights}
\label{sec:analysis_weights}

To further validate the effectiveness of our weighted news aggregation, we analyze the distribution of generated news weights and use case studies for a comprehensive understanding. Since we transform news sentiments into trainable representations, which cannot be directly compared to the original sentiment scores, we focus on individual news utilization to demonstrate how we mitigate sentiment homogenization issues.

\begin{figure}[b]
\vspace{-0.8cm}
\centering
\begin{center}
   \captionof{figure}{The kernel density esimate plot of MANA-Net's news weights on the test set.}
   \includegraphics[width=0.98\linewidth]{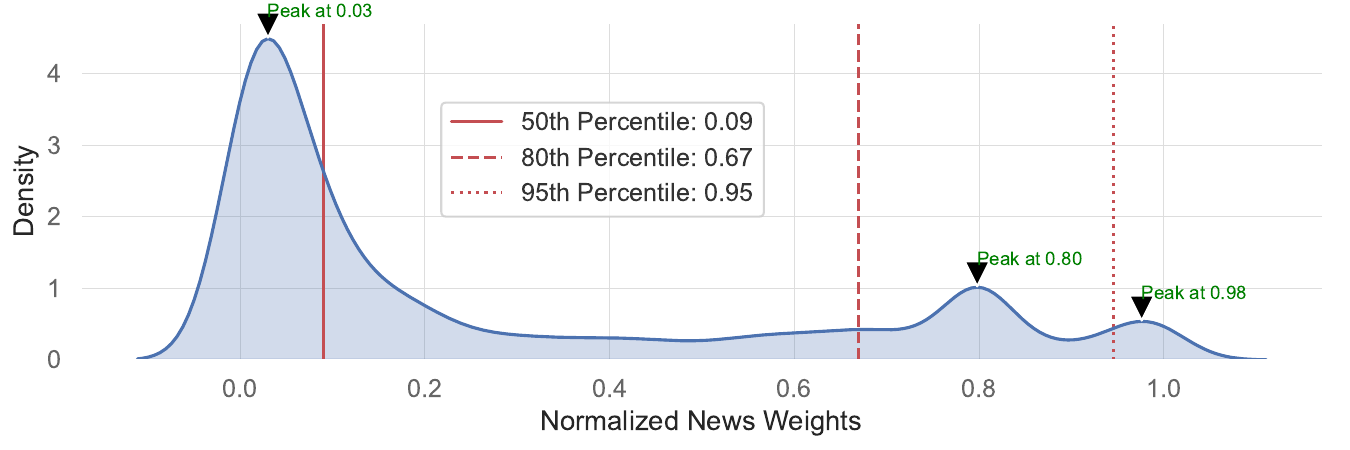}
\label{fig:score_distribution}
\end{center}
\vspace{-0.1cm}
\end{figure}

Figure~\ref{fig:score_distribution} depicts the distribution of news weights across the test set. Since the range of news weights varies with daily news volume, we normalize these weights to a 0-1 scale, maintaining their rankings and ensuring comparability across different days. A key observation is that news weights are not uniform, reflecting varying degrees of market influence. Notably, the percentile lines indicate that over 50\% of news items are assigned very low weights, suggesting that most news has only a minimal impact on market movements. Furthermore, more than 80\% of news items receive weights below 0.8, illustrating that the majority do not significantly stand out amidst the voluminous news data. 
Conversely, fewer than 5\% of news items have weights exceeding 0.98, highlighting a select few pieces of news that are particularly influential and warrant specific attention. This distribution confirms that MANA-Net assigns varying weights to news items, emphasizing certain news by allocating higher weights, thereby distinguishing their relative importance in influencing market movements.

To illustrate the functionality of MANA-Net's news weighting system, we analyze several news cases. Analyzing every news item is infeasible due to the sheer volume of daily news and the intricate link between news and market fluctuations. While many news pieces, like ``Johnson \& Johnson Completes Divestiture of LifeScan to Platinum Equity'', provide useful information, their specific market impact is difficult for humans to ascertain. Therefore, for a clear and focused analysis, we select major news items that apparently exert significant market influence. Table~\ref{tab:case_study} lists these selected news cases, including broad market news such as the performance of the Dow Industrials and statements by Macron affecting EU and global markets. We also include influential incidents with clear market consequences,  the 2018 murder of Saudi journalist Khashoggi, which significantly affected U.S.-Saudi relations and market sentiment.

These selected news items are recognized for their significant influence on the markets. In MANA-Net, each of them received a weight above 0.98, surpassing over 95\% of all news items analyzed. The high weights underscores that MANA-Net effectively elevates the importance of these influential news items in its predictions. Furthermore, the sentiment scores of these news items are exceptionally polarized, with values exceeding 0.8, indicating strong positive or negative impacts. However, the last column shows that the corresponding daily average sentiments are highly homogenized, obscuring the distinct sentiments of these influential news items. This discrepancy highlights the shortcomings of equal-weight aggregation methods and demonstrates how MANA-Net successfully leverages critical news information to enhance market prediction.

\vspace{-0.2cm}
\section{Conclusion}
In this paper, we address a significant challenge existing in financial news sentiment aggregation: Aggregated Sentiment Homogenization. This issue was explored through the analysis of an extensive news dataset. We developed MANA-Net, which is specifically designed to counteract this challenge. MANA-Net employs a market-news attention mechanism that dynamically aggregates news data, assigning varying weights to news items based on their assessed market impact. This approach not only facilitates a more sophisticated utilization of news sentiments but also ensures that the aggregation process remains adaptable and trainable, thus producing more effective news representations for market prediction. Experimental results underscore MANA-Net's superior performance relative to various contemporary methods, highlighting its successful exploitation of news information for market analysis.

\bibliographystyle{ACM-Reference-Format}
\bibliography{ref}

\end{document}